\documentclass[10pt,twocolumn,letterpaper]{article}

\usepackage[pagenumbers]{cvpr} 
\usepackage{multirow} 
\definecolor{cvprblue}{rgb}{0.21,0.49,0.74}
\usepackage[pagebackref,breaklinks,colorlinks,allcolors=cvprblue]{hyperref}

\def\paperID{2717} 
\def\confName{CVPR}
\def\confYear{2025}

\title{Discrete to Continuous: Generating Smooth Transition Poses\\ from Sign Language Observations}

\author{
Shengeng Tang~~~~~
Jiayi He~~~~~
Lechao Cheng$^*$~~~~~
Jingjing Wu~~~~~
Dan Guo~~~~~
Richang Hong$^*$\\
\textsuperscript{}School of Computer Science and Information Engineering, Hefei University of Technology \\
{\tt\small \{tangsg, chenglc, wujingjing, guodan, hongrc\}@hfut.edu.cn, 2024170906@mail.hfut.edu.cn}
}

\begin{document}
\maketitle
\begin{abstract}

Generating continuous sign language videos from discrete segments is challenging due to the need for smooth transitions that preserve natural flow and meaning. Traditional approaches that simply concatenate isolated signs often result in abrupt transitions, disrupting video coherence. To address this, we propose a novel framework, Sign-D2C, that employs a conditional diffusion model to synthesize contextually smooth transition frames, enabling the seamless construction of continuous sign language sequences. Our approach transforms the unsupervised problem of transition frame generation into a supervised training task by simulating the absence of transition frames through random masking of segments in long-duration sign videos. The model learns to predict these masked frames by denoising Gaussian noise, conditioned on the surrounding sign observations, allowing it to handle complex, unstructured transitions. During inference, we apply a linearly interpolating padding strategy that initializes missing frames through interpolation between boundary frames, providing a stable foundation for iterative refinement by the diffusion model. Extensive experiments on the PHOENIX14T, USTC-CSL100, and USTC-SLR500 datasets demonstrate the effectiveness of our method in producing continuous, natural sign language videos. 
\end{abstract} 
\section{Introduction}
\label{sec:intro}
Sign language~\cite{tang2021graph, Koller_2020, 10109128, zhao2023best, hu2022collaborative, guo2021sign, camgoz2020multi, gong2024llms}, as a unique form of visual communication, presents an intricate challenge for data collection, especially when scaling datasets for continuous, long-duration sign sentences. Unlike textual language, which can be assembled seamlessly from words to sentences, sign language requires a fluid connection between individual signs to convey meaning accurately. This connection involves not only the specific signs but also the transitions between them, which convey temporal and contextual coherence vital to understanding the intended message. As a result, while word-level sign language data~\cite{dreuw2008benchmark, zhang2016chinese, chai2014devisign, li2020word} has been collected extensively, there remains a critical gap in developing robust, large-scale datasets for sentence-level sign language~\cite{huang2018video, duarte2019cross, Camgoz_Hadfield_Koller_Ney_Bowden_2018}, particularly for continuous sequences. 

\begin{figure}[t]
\centering
\includegraphics[width=\linewidth]{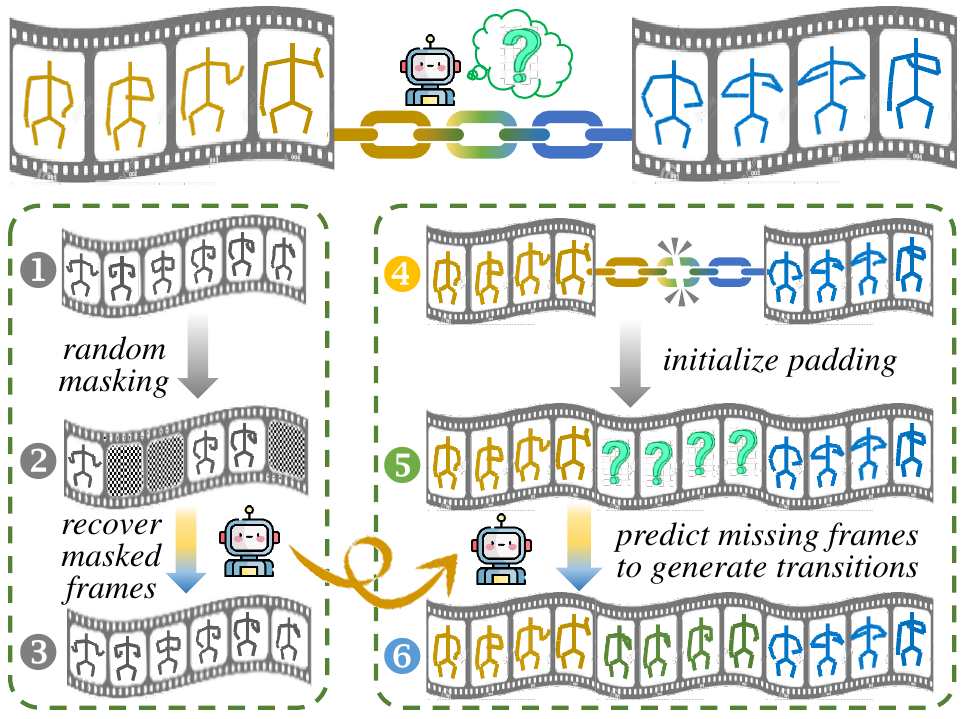}
\caption{Task and key steps. Our work aims to generate continuous sign videos by creating transition poses between discrete segments. In training, random masking simulates missing transitions, and the model learns to recover these frames (steps 1-3). During inference, padding initializes missing transitions, which the model refines to generate smooth, coherent sequences (steps 4-6).}
\label{fig:teaser}
\end{figure}

A possible workaround involves pre-collecting word-level sign data akin to a dictionary~\cite{joy2020developing, de2023querying, aliwy2021development} and then attempting to form sentences by joining these discrete signs. However, assembling sign sentences is fundamentally different from assembling textual sentences due to the need for visually smooth transitions between signs, which maintain natural flow and meaning. Direct concatenation of individual signs into sentences often results in abrupt and unnatural transitions, disrupting the continuity of the video and potentially obscuring meaning. Consequently, generating smooth transition frames between sign segments is essential to synthesizing continuous sign language videos~\cite{Zelinka_Kanis_2020, sugandhi2020sign, zeng2020highly}, yet presents significant challenges due to the lack of supervised reference data for the missing transition frames. 

To address this issue, we propose a smooth transition pose generation framework, named Sign-D2C, to synthesize \textbf{Continuous} sign language videos from \textbf{Discrete} sign segments. This framework is based on a diffusion model, which learns to generate missing frames by observing surrounding sign frames. Specifically, our method leverages a conditional diffusion model trained to predict the missing transition frames by sampling from Gaussian noise and being conditioned by the observed video frames. As illustrated in Fig.~\ref{fig:teaser}, our method consists of a two-phase process designed to overcome the limitations of direct sign concatenation. \textbf{In the training phase}, we simulate the transition frame generation task by randomly masking frames within long sign language videos. This random masking effectively removes video segments, mimicking real-world scenarios where transition frames are absent. The model is then trained to reconstruct these missing frames based on the observable surrounding frames. By doing so, the model learns to generate plausible transitions in a supervised manner, enhancing its robustness and adaptability to complex, unstructured transitions. 
\textbf{In the inference phase}, we introduce a linearly interpolating padding strategy to initialize the missing transition frames. This strategy begins by selecting the last frame of the preceding observed segment and the first frame of the following observed segment as the boundary frames for the missing transition segment. The intermediate frames within the missing segment are then initialized by linearly interpolating between these boundary frames, creating an initial linearly interpolating padding that provides a stable foundation for generating smooth transitions. The conditional diffusion model then iteratively refines these frames to produce naturalistic transition frames that connect the observed segments seamlessly. The contributions of this work are summarized as follows: 

\begin{itemize}
    \item We propose a novel conditional diffusion framework, Sign-D2C, specifically designed to synthesize smooth and contextually accurate transition frames for continuous sign videos, overcoming the limitations of traditional word-to-sentence assembly techniques in sign language. 
    \item To simulate the generation of transition frames, we introduce a random masking strategy that removes sections of long sign language videos during training, enabling the model to predict missing frames in a supervised manner. This approach enhances the model’s ability to handle diverse, complex transitions. 
    \item We design a linearly interpolating padding strategy for transition initialization in the inference phase. This strategy utilizes the boundary frames of observed segments to initialize the missing poses with linear interpolation, which provides a stable basis for generating visually seamless transitions.  
    \item Extensive experiments on three standard benchmarks (\emph{i.e.}, Phoenix14T, UCST-CSL100, and USTC-SLR500) demonstrate the robustness and effectiveness of our approach in producing smooth, contextually accurate transition poses for continuous sign language videos. 
\end{itemize}

\section{Related Work}
\label{sec:related}
\subsection{Sign Language Production \& Synthesis}
Sign language plays a vital role in daily life and has become a prominent topic in artificial intelligence. Early research primarily focused on Sign Language Recognition (SLR)~\cite{Koller_2020, 10109128, zhao2023best, hu2022collaborative, Guo_Zhou_Li_Wang_2018} and Sign Language Translation (SLT)~\cite{Guo_Tang_Wang, camgoz2020multi, li2020tspnet, gong2024llms}. However, with the growing need for effective communication between deaf and hearing individuals, Sign Language Production (SLP) has gained significant attention in recent years. 

Early approaches to SLP utilize synthetic animation techniques~\cite{Mazumder_Mukhopadhyay_Namboodiri_Jawahar_2021, McDonald_Wolfe_Schnepp_Hochgesang_Jamrozik_Stumbo_Berke_Bialek_Thomas_2016, Segouat_2009} to convert sentences into sign animations. These techniques, however, incur high collection costs and are limited to specific terms. With advancements in deep learning models, SLP has seen more extensive exploration. Stoll \emph{et al.}~\cite{Stoll_Camgoz_Hadfield_Bowden_2020} proposes a multi-step approach for generating sign language videos from text, involving the transformation of text into gloss, gloss into poses, and finally, poses into video. Building on this, Saunders \emph{et al.}~\cite{saunders2020progressive} introduces a model that generates sign language poses through autoregression. Diffusion-based solutions like G2P-DDM~\cite{xie2024g2p} and GCDM~\cite{tang2024GCDM} further advance SLP by generating sign poses under semantic guidance. 

Despite significant progress in these methods, the limitations of existing data have become a major bottleneck in the development of sign language technology. Previous Sign Language Synthesis (SLS) approaches~\cite{Zelinka_Kanis_2020} focus on extracting gestures from continuous sign language videos. However, these methods still face challenges in synthesizing continuous, long-duration sign language sentences. Our method addresses this gap by starting with word-level sign language pose data, generating smooth transition frames between individual signs, and combining these segments into sentence-level sequences, thereby overcoming the limitations of existing methods. 

\subsection{Diffusion-based Generative Models}

Diffusion models have gained considerable attention in generative modeling since they were first introduced by Sohl-Dickstein \emph{et al.}\cite{Sohl-Dickstein_Weiss_Maheswaranathan_Ganguli_2015}. Its forward-backward diffusion process has proven highly effective in generating realistic and diverse outputs, enabling diffusion models to achieve competitive performance in various generative tasks, including image synthesis~\cite{rombach2022high}, video generation\cite{yang2024fresco, lee2024grid}, and 3D shape generation~\cite{mo2023dit}. 
One of the strengths of diffusion models is their iterative refinement capability, allowing for flexible and high-quality outputs that align well with tasks requiring fine control over generative processes. This quality has led to their adoption in areas such as image restoration, where they have been used for denoising and super-resolution~\cite{miao2023dds2m, xia2023diffir, yi2023diff}, as well as pose estimation, where they can handle uncertainty in human poses~\cite{shan2023diffusion}. Recent advancements have extended their use to more complex generative tasks, such as controlled text-to-image synthesis~\cite{xie2023boxdiff} and conditional video generation~\cite{ni2023conditional}, further demonstrating their adaptability and robustness.

In Sign Language Production (SLP), diffusion models remain relatively unexplored but have shown promise in recent studies. For instance, G2P-DDM~\cite{xie2024g2p} uses a discrete diffusion model to translate the continuous pose generation task into a discrete space, simplifying the complexity of generating sign language poses while maintaining semantic alignment with gloss-based inputs. This approach allows for finer control over pose accuracy, making it particularly suitable for generating static sign poses from linguistic cues. Similarly, GCDM~\cite{tang2024GCDM} applies a multi-hypothesis diffusion framework to generate diverse sign poses from gloss, accommodating multiple interpretations and improving the naturalness of synthesized gestures. These models represent pioneering steps in applying diffusion-based approaches to SLP, leveraging the unique properties of diffusion models to generate realistic sign poses under semantic guidance.

Despite these advancements, current diffusion-based methods for SLP are primarily focused on generating sign poses according to textual signals rather than addressing the temporal continuity required for longer sign language sequences. In sign language synthesis, especially when producing smooth transitions in continuous sign language videos, maintaining smooth transitions between gestures is essential for ensuring that the sequences are fluid and natural. This need for temporally smooth transitions has led to research on methods that can generate these connecting frames between segments, although it remains an area with limited exploration in SLP. 
Beyond SLP, diffusion models have been successfully applied to tasks requiring temporal coherence and frame interpolation in video synthesis. For example, diffusion-based models have been employed in video generation to ensure smooth transitions between frames~\cite{liang2025movideo}, and similar techniques have been explored in action synthesis and motion completion~\cite{karunratanakul2023guided}, where maintaining continuity across frames is crucial. Such applications highlight the potential of diffusion models to handle tasks involving complex temporal dependencies, suggesting that similar techniques could be valuable in SLP for generating continuous sign language videos.

\section{Methodology}
\label{sec:method}

\begin{figure*}[t]
\centering
\includegraphics[width=\textwidth]{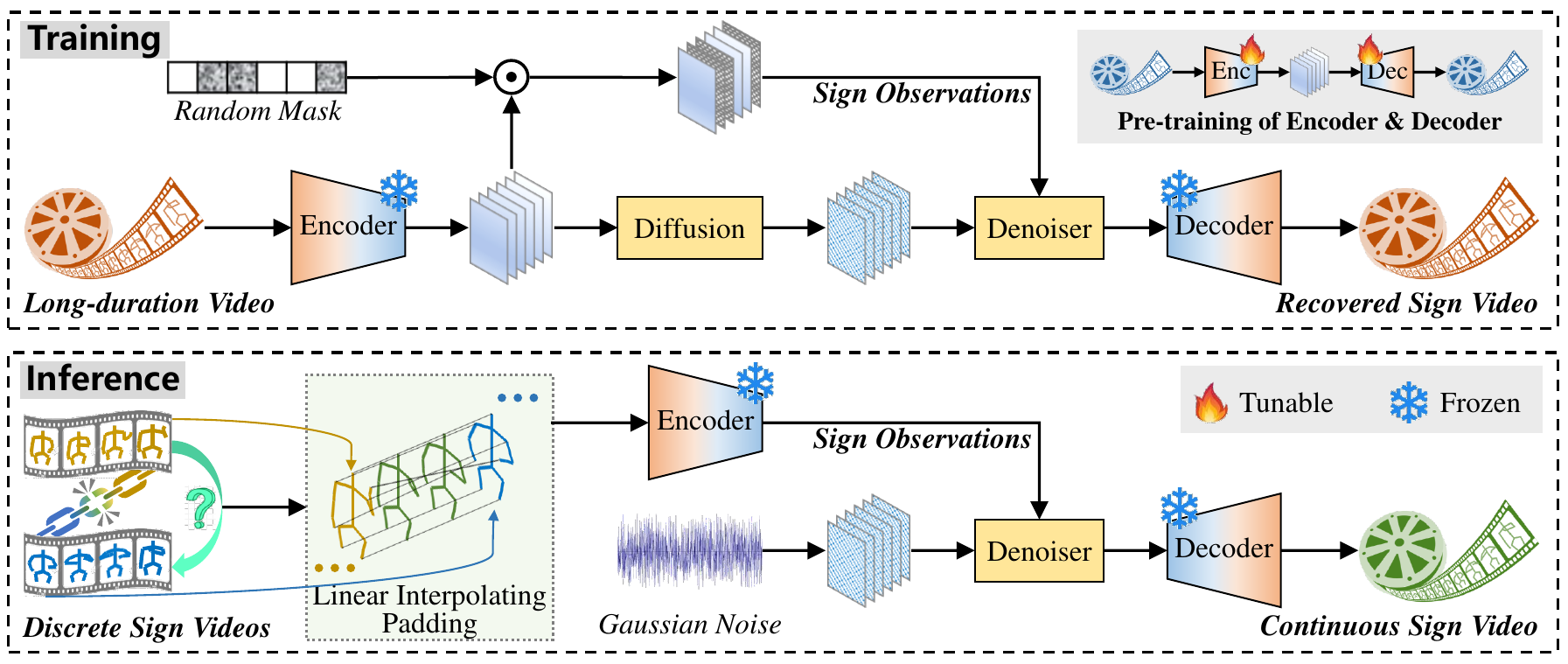}
\caption{Overview. The proposed framework for generating continuous sign language videos with smooth transitions between discrete segments. In the training phase (top), a long-duration sign video undergoes random masking to create gaps, simulating missing transitions. An encoder processes the video, and a conditional diffusion model denoises and predicts the masked segments based on observable frames. The recovered video is produced through a decoder, with both encoder and decoder pre-trained for effective representation. In the inference phase (bottom), discrete sign videos are connected using a linearly interpolating padding strategy to initialize missing transitions. Gaussian noise is applied, and the model refines these transitions, generating a continuous sign video with coherent flow.}
\label{fig: overview}
\end{figure*}

In this work, we propose a framework for generating smooth transition frames between discrete sign segments to create continuous sign pose videos. Our approach transforms an unsupervised transition generation task into a supervised framework using a random masking strategy based on conditional diffusion. This section will outline the preliminaries (Sec.~\ref{subsec:preliminary}), followed by our core idea of realizing the task (Sec.~\ref{subsec:idea}), the framework structure (Sec.~\ref{subsec:framework}), and the training process (Sec.~\ref{subsec:training}).

\subsection{Preliminary}
\label{subsec:preliminary}
The diffusion model~\cite{Ho_Jain_Abbeel_Berkeley} is highly effective in generating high-quality images by combining diffusion processes, latent space modeling, and conditional inputs. In this work, we employ a conditional diffusion model to synthesize smooth transition frames between discrete sign segments, forming continuous sign language videos. Here, the model is trained on complete long-duration sign language pose sequences, progressively adding noise to simulate missing transitions. During testing, the model starts from Gaussian noise samples and generates the required transitions between observed sign frames.

In the forward process, noise is incrementally added to the full sign language pose sequence $x_0$ to obtain a noisy version $x_T$,  mimicking the degradation of visual continuity. This process is described by: 
\begin{eqnarray}
    x_t = \alpha_t \cdot x_{t-1} + \sqrt{1 - \alpha_t} \cdot \epsilon_t,
    \label{eq: noise add}
\end{eqnarray}
where $\alpha_t$ controls the level of noise at each step $t$, and $\epsilon_t$ represents Gaussian noise. As $t$ increases, the original sequence $x_0$ transitions to pure noise $x_T$.

The reverse process reconstructs the continuous sequence by progressively removing noise, where a neural network is trained to predict each denoised frame based on previous noisy frames. The denoising step at each iteration is defined as: 
\begin{eqnarray}
p(x_{t-1} | x_t, x_{\text{obs}}) = \mathcal{N}(x_{t-1}; \mu_\theta(x_t, t, x_{\text{obs}}), \Sigma_t),
\end{eqnarray}
where $x_{\text{obs}}$ denotes the observed sign language frames that serve as conditional inputs, and $\mu_\theta(x_t, t, x_{\text{obs}})$ is the predicted mean for the denoising process. The observed frames condition the diffusion model, guiding it to produce smooth, contextually consistent transition frames.

In our model, the training input is a complete long-duration sign language pose video, gradually noised to simulate missing transitions. During testing, Gaussian noise serves as the input, and the model generates a continuous video with smooth transition frames conditioned on observed frames $x_{\text{obs}}$, effectively bridging discrete sign segments into a fluid sign language sequence. 



\subsection{Core Idea: Unsupervised to Supervised}
\label{subsec:idea}
Generating continuous sign language videos by adding transitions between discrete segments is challenging, as these segments typically start and end with specific gestures that make direct concatenation impractical. Additionally, this transition generation task is inherently unsupervised, as real transition frames are generally unavailable for training. To address this, we convert the \textbf{unsupervised} problem of generating transitions into a \textbf{supervised} learning framework by leveraging existing continuous sign language videos.

Given a complete continuous sign language sequence $X=\{x_1, x_2,\cdots, x_n \}$, we simulate missing transitions by randomly masking portions of $X$, creating gaps that mimic real-world transition absences. Let the masked frames be denoted as $M \subset X$, and the observable pose frames as $X_\text{obs} = (X \setminus X_\text{mask})$. This setup frames the task as predicting $X_\text{mask}$ based on $X_\text{obs}$, effectively generating supervised training data for transition prediction. 

By training the model to predict the masked frames $X_\text{mask}$ conditioned on the observable context $X_\text{obs}$, it learns to generate realistic transitions that smoothly bridge the gap. This innovative masking strategy allows our model to leverage the context of surrounding frames provided by $X_\text{obs}$ to produce temporally coherent connections, transforming the inherently unsupervised transition generation task into a supervised one and enabling effective continuous sign language synthesis. 


\subsection{Sign-D2C Framework}
\label{subsec:framework}
Our approach leverages a conditional diffusion model to generate transition frames, enabling smooth continuity between discrete sign segments. The model architecture consists of three main components: an encoder, a decoder, and a denoiser. The first two are composed of two self-attention layers and the denoiser consists of two blocks, each containing a self-attention layer, a cross-attention layer, and a feed-forward layer. This setup allows the model to effectively capture temporal dependencies and complex contextual information needed for sign language synthesis.

The Encoder first processes the input sequence $X = \{x_1, x_2, \ldots, x_n\}$, where $x_i$ represents individual sign language pose frames. By encoding the long-duration video input, the Encoder transforms $X$ into a latent representation $Z$, which serves as the foundation for both the diffusion and denoising processes: 
\begin{eqnarray}
Z = \text{Encoder}(X).
\end{eqnarray}

During the forward diffusion, Gaussian noise is gradually added to $Z$ at each time step $t$, resulting in a noisy representation $Z_t$. This process can be transferred from Eq.~\ref{eq: noise add}, formulated as: 
\begin{eqnarray}
Z_t = \alpha_t \cdot Z_{t-1} + \sqrt{1 - \alpha_t^2} \cdot \epsilon_t,
\end{eqnarray}
where $\alpha_t$ controls the noise schedule, and $\epsilon_t$ is sampled from a Gaussian distribution. The diffusion process progressively transforms the latent representation towards pure noise.

The denoiser then reconstructs the original representation by iteratively removing the noise. Here, the denoiser is conditioned on sign observations $X_\text{obs}$, which serve as contextual priors. These observations are incorporated into the denoiser via cross-attention, allowing the model to focus on the relevant parts of $X_\text{obs}$ at each denoising step. For each time step $t$ in the reverse process, the denoised output $\hat{Z}_{t-1}$ is given by: 
\begin{eqnarray}
\hat{Z}_{t-1} = \text{Denoiser}(Z_t, X_\text{obs}),
\label{eq:denoiser}
\end{eqnarray}
where $X_\text{obs}$ provides the model with the necessary context to predict the masked segments accurately. 

Finally, the decoder, also structured with three self-attention layers, takes the denoised latent representation and transforms it back into the original sign language video space, outputting the recovered or synthesized video $\hat{X}$: 
\begin{eqnarray}
\hat{X} = \text{Decoder}(\hat{Z}_0).
\end{eqnarray}

By using self-attention within both the encoder and decoder, and cross-attention in the denoiser to incorporate contextual observations, the model effectively learns to generate temporally coherent transitions, resulting in a smooth, continuous sign language video output. 

The proposed Sign-D2C incorporates two key strategies to enhance transition generation: a random masking strategy during training and linearly interpolating padding initialization for inference. The following provides details. 

\noindent\textbf{Random Masking Strategy.}
To generate smooth transitional poses and capture relationships between different segments, we employ a random masking strategy during training, as illustrated in the training phase of Fig.~\ref{fig: overview}. 

Given a long-duration video $p\in\mathbb{R}^{F\times (N\times3)}$ with $F$ frames, each containing $N$ joints in 3D space, the video is first encoded by a pre-trained encoder into a latent representation $x \in \mathbb{R}^{F\times W}$, where $W$ denotes the dimension of the latent feature $x$. The mask $m$ is constructed by randomly selecting frame indices, extending across the dimensions of the latent feature $x$. This mask $m$ is then applied to selectively retain or mask out parts of $x$, resulting in a masked latent representation: 
\begin{eqnarray}
    x^m = x \odot m, x \in \mathbb{R}^{F\times W}, 
\end{eqnarray}
where $x^m$ serves as a conditional input to the model, distinguishing between occluded and visible movements. This masking strategy enables the model to predict missing frames based on observable context, facilitating smooth transition generation.

\noindent\textbf{Linear Interpolation Padding.}
In the inference phase, generating smooth transitions between discrete motion segments is challenging due to the gaps between observed frames. We leverage the natural correlation between consecutive frames in sign language sequences by introducing a linear interpolation padding initialization strategy, as illustrated in Fig.~\ref{fig: overview}. 

This strategy starts by using the last pose of the preceding observed segment, $p_s$, and the first pose of the following observed segment, $p_e$, as boundary frames for the missing transition segment. The goal is to generate intermediate frames $p_c$ that provide a smooth transition from $p_s$ to $p_e$. Using linear interpolation, we initialize these missing frames as follows: 
\begin{eqnarray}
\left\{\begin{array}{l}
    f(i) = \sum_{k=0}^{i} \frac{1}{(k + 1)}, \quad i = 0, \cdots, (e - s),\\
    p_c(i) = p_s + \frac{p_e - p_s}{e - s + 1} \cdot f(i).
\end{array}\right.
\label{eq:padding}
\end{eqnarray}

This interpolation scheme gradually fills in the key points at each time step, creating intermediate poses that ensure a coherent and natural flow between segments. This strategy facilitates smoother transitions and provides a stable foundation for later refinement, resulting in seamless sign language video synthesis. 

\subsection{Training}
\label{subsec:training}
\noindent\textbf{Pre-training of Encoder \& decoder.}
To enable coherent video generation, we pre-train the encoder and decoder to ensure effective latent representation processing. The encoder extracts latent features $x$ from input videos, while the decoder reconstructs realistic sign language poses from these latent representations. For training the diffusion model, we freeze the encoder and decoder parameters, allowing the model to focus on learning smooth transitions based on these pre-trained features. 

Pre-training is conducted end-to-end with a reconstruction loss, optimizing the encoder-decoder pair to accurately represent and recover video details: 
\begin{eqnarray} 
\mathcal L_\text{pre} = \frac{1}{F}\sum_{f=1}^{F}|p_f - \hat{p}_f|, 
\end{eqnarray}
where $p$ represents the input video with $F$ frames, and $\hat{p}$ is the reconstructed output. This process ensures the encoder captures essential features, while the decoder faithfully reconstructs realistic sign language poses, setting a solid foundation for high-quality transition generation. 

\noindent\textbf{Loss Optimization.} 
According to Eq.~\ref{eq:denoiser}, we can obtain the denoiser prediction $\hat{X}=\{ \hat{x}_1, \hat{x}_2, \cdots, \hat{x}_F\}$. During training, the loss function constrains the difference between the original latent feature $x$ and  the denoiser’s prediction $\hat{x}$: 
\begin{eqnarray} 
\mathcal L_\text{diff} = \frac{1}{F}\sum_{f=1}^{F}|x_f-\hat{x}_f|,
\end{eqnarray}
where $F$ denotes the number of frames in the video sequence. This loss encourages the model to accurately recover the original latent representation, improving the quality of the generated transitions and ensuring alignment with the encoded video features. 



\section{Experiments}
\label{sec:experiments}
\subsection{Experimental Settings}
\noindent\textbf{Datasets.} 
We conduct the experiments on three datasets, \emph{i.e.}, PHOENIX14T, USTC-CSL100, and USTC-SLR500. The first two are sentence-level continuous sign language datasets, which are used to evaluate supervised transition generation in controlled settings. While the latter one is a word-level isolated sign language dataset, allowing us to assess the model's performance in generating transitions within real-world, discrete sign scenarios.
\begin{itemize}
\item \textbf{PHOENIX14T}~\cite{Camgoz_Hadfield_Koller_Ney_Bowden_2018} comprises 8,257 sentence-level samples, which covers 2,887 German words. For preprocessing, we follow the settings in previous sign language production studies~\cite{huang2021towards, saunders2021mixed}, using OpenPose~\cite{cao2017realtime} to extract 2D coordinates from the videos, which are then refined to 3D coordinates through a skeleton correction model~\cite{Zelinka_Kanis_2020}. 
\item \textbf{USTC-CSL100}~\cite{huang2018video} covers 100 unique sentences performed by 50 signers. We use the split-I~\cite{guo2018hierarchical} approach to divide the dataset into 4,000 training examples and 1,000 test examples. The above two datasets enable us to evaluate the model’s effectiveness on continuous sign language sequences in a structured environment.
\item \textbf{USTC-SLR500}~\cite{huang2018attention} is an isolated sign language dataset containing 500 Chinese sign words, each performed five times by 50 signers. Following~\cite{huang2018attention}, we split the dataset by using videos from 36 signers for training and the remaining 14 signers for testing. This setup enables us to test the model's ability to generate transitions between isolated signs in a realistic, word-level scenario.
\end{itemize}

\noindent\textbf{Implementation Details.} 
Since generating continuous sign language videos from discrete fragments is inherently an unsupervised task, there are no standardized evaluation metrics. To effectively assess our model’s performance, we analyzed transition actions between two gestures on the sentence-level datasets (\emph{i.e.}, PHOENIX14T and USTC-CSL100), observing that transitions typically span 10 to 20 frames. Based on these findings, we designed experiments using two masking strategies: masking 10 frames every 20 frames and masking 20 frames every 10 frames. 
Additionally, to further validate the effectiveness of our approach, we used the isolated sign language dataset (\emph{i.e.}, USTC-SLR500) to assemble discrete vocabulary videos into continuous sequences, demonstrating the model’s capability to produce smooth transitions in a real-world setting.

\noindent\textbf{Evaluation Metrics.}
We evaluate our performance on two sentence-level datasets in terms of both semantics and coherence. Following prior work in sign language generation~\cite{huang2021towards, saunders2020progressive, saunders2021mixed, Viegas_Inan_Quandt_Alikhani}, we use the NSLT method~\cite{Camgoz_Hadfield_Koller_Ney_Bowden_2018} to convert generated sign language videos back into text. We then compare this text with reference data to calculate semantic accuracy metrics, including BLEU~\cite{papineni2002bleu}, ROUGE~\cite{lin2004rouge}, and WER. Additionally, we use FID, DTW, and MPJPE to assess the consistency and smoothness of the generated transitions, measuring the coherence of the results. 

\begin{table}[tbp]
\renewcommand\arraystretch{1.1}
\caption{Ablation results of different masking ratios on PHOENIX14T (observation 20 frames - prediction 10 frames).}
\vspace{-3mm}
\label{tab: random mask ratio 20-10}
\centering
\resizebox{\linewidth}{!}{
\begin{tabular}{cccccccc}
\toprule[1pt]
   \multirow{2}{*}{Masking Ratio $r$} & \multicolumn{3}{c}{DEV} & ~ & \multicolumn{3}{c}{TEST} \\
   \cline{2-4}\cline{6-8}
    & B1{$\uparrow$} & B4{$\uparrow$} & DTW{$\downarrow$} & ~ & B1{$\uparrow$} & B4{$\uparrow$} & DTW{$\downarrow$}\\
\toprule[0.5pt]
   $r$ = 0.1 &22.17 &8.04 &\bf{0.61} & ~ &\bf{22.47} &8.67 &\bf{0.60}\\
   $r$ = 0.3 & \bf{22.46} & \bf{8.06} & 0.62 & ~ & 22.36 & \bf{8.73} & 0.61\\
   $r$ = 0.5 &22.30 &\bf{8.06} &0.77 & ~ &22.02 &8.35 &0.75 \\
\toprule[1pt]
\end{tabular}}
\vspace{-3mm}
\end{table}



\begin{table}[tbp]
\renewcommand\arraystretch{1.1}
\caption{Ablation results of different padding strategies on PHOENIX14T (observation 20 frames - prediction 10 frames).}
\vspace{-3mm}
\label{tab: ablation padding}
\centering
\resizebox{\linewidth}{!}{
\begin{tabular}{cccccccc}
\toprule[1pt]
   \multirow{2}{*}{Padding Strategy} & \multicolumn{3}{c}{DEV} & ~ & \multicolumn{3}{c}{TEST} \\
   \cline{2-4}\cline{6-8}
    & DTW{$\downarrow$} & FID{$\downarrow$} & MPJPE{$\downarrow$} & ~ & DTW{$\downarrow$} & FID{$\downarrow$} & MPJPE{$\downarrow$}\\
\toprule[0.5pt]
   w/o Padding    &11.46 &2.74 &69.60 & ~ &11.60 &3.06 &87.57 \\
   Front Padding  &0.72 &2.68 &47.90  & ~ &0.69 &2.99 &60.05 \\
   Linear Padding  & \bf{0.62} & \bf{2.68} & \bf{47.30} & ~ & \bf{0.61} & \bf{2.98} & \bf{59.36}\\
   Back Padding  &0.67 &2.68 &47.89  & ~ &0.66 &2.99 &59.91 \\
\toprule[1pt]
\end{tabular}}
\vspace{-3mm}
\end{table}


\begin{table*}[!htbp]
\renewcommand\arraystretch{1.1}   
\caption{Performance comparison on PHOENIX14T (under setting: observation 20 frames - prediction 10 frames).}
\vspace{-3mm}
\centering
\resizebox{\textwidth}{!}{
\begin{tabular}{lccccccccccccccc}
\toprule[1pt]
\multirow{2}{*}{Methods}&\multicolumn{7}{c}{DEV} & ~ &\multicolumn{7}{c}{TEST} \\
\cline{2-8}
\cline{10-16}
~ &B1$\uparrow$ &B4$\uparrow$ &ROUGE$\uparrow$ &WER$\downarrow$ &DTW$\downarrow$ &FID$\downarrow$ &MPJPE$\downarrow$& ~ &B1$\uparrow$ &B4$\uparrow$ &ROUGE$\uparrow$ &WER$\downarrow$ &DTW$\downarrow$ &FID$\downarrow$ &MPJPE$\downarrow$\\ 
\midrule[0.5pt]
\multicolumn{1}{l}{Ground Truth}  &29.77 &12.13 &29.60 &74.17&0.00&0.00&0.00 &~ &29.76 &11.93 &28.98 &71.94 &0.00&0.00&0.00\\ 
\midrule[0.5pt]
\multicolumn{1}{l}{G2P-DDM~\cite{xie2024g2p}} &8.25 &0.99 &7.17 &98.93 &12.71 &2.85 &55.02 &~ &9.52 &1.29 &8.12 &98.54 &12.57 &3.19 &67.81 \\
\multicolumn{1}{l}{VQ-GCDM~\cite{tang2024GCDM}} &9.31 &0.87 &7.59 &98.69 &12.11 &2.86 &53.87 & ~ &9.85 &0.96 &7.95 &98.59 &11.98 &3.21 &66.51 \\
\midrule[0.5pt]
\multicolumn{1}{l}{\textbf{Ours}}
&\bf{22.46}&\bf{8.06}&\bf{22.07} &\bf{80.73}&\bf{0.62} &\bf{2.68} &\bf{47.30}& ~ &\bf{22.36}&\bf{8.73}&\bf{22.36}&\bf{79.06} &\bf{0.61} &\bf{2.98} &\bf{59.36}\\ 
\bottomrule[1pt]
\end{tabular}}
\label{tab:main1-1}
\vspace{-3mm}
\end{table*}

\begin{table*}[!htbp]
\renewcommand\arraystretch{1.1}   
\caption{Performance comparison on PHOENIX14T (under setting: observation 10 frames - prediction 20 frames).}
\vspace{-3mm}
\centering
\resizebox{\textwidth}{!}{
\begin{tabular}{lccccccccccccccc}
\toprule[1pt]
\multirow{2}{*}{Methods}&\multicolumn{7}{c}{DEV} & ~ &\multicolumn{7}{c}{TEST} \\
\cline{2-8}
\cline{10-16}
~ &B1$\uparrow$ &B4$\uparrow$ &ROUGE$\uparrow$ &WER$\downarrow$ &DTW$\downarrow$ &FID$\downarrow$ &MPJPE$\downarrow$& ~ &B1$\uparrow$ &B4$\uparrow$ &ROUGE$\uparrow$ &WER$\downarrow$ &DTW$\downarrow$ &FID$\downarrow$ &MPJPE$\downarrow$\\ 
\midrule[0.5pt]
\multicolumn{1}{l}{Ground Truth}  &29.77 &12.13 &29.60 &74.17&0.00&0.00&0.00 &~ &29.76 &11.93 &28.98 &71.94 &0.00&0.00&0.00\\ 
\midrule[0.5pt]
\multicolumn{1}{l}{G2P-DDM~\cite{xie2024g2p}} &8.98 &0.63 &7.44 &98.61 &15.95 &2.89 &55.21 &~ &9.13 &0.74 &7.75 &98.64 &15.58 &3.22 &68.18 \\
\multicolumn{1}{l}{VQ-GCDM~\cite{tang2024GCDM}} &8.64 &0.89 &7.08 &98.48 &12.40 &3.05 &48.66 & ~ &9.02 &1.36 &7.22 &98.61 &12.22 &3.37 &60.27 \\
\midrule[0.5pt]
\multicolumn{1}{l}{\textbf{Ours}}
&\bf{19.32}&\bf{5.92}&\bf{18.33} &\bf{87.30}&\bf{0.70} &\bf{2.75} &\bf{48.04}& ~ &\bf{19.59}&\bf{6.26}&\bf{18.59}&\bf{86.8} &\bf{0.68} &\bf{3.05} &\bf{59.81}\\ 
\bottomrule[1pt]
\end{tabular}}
\label{tab:main1}
\vspace{-3mm}
\end{table*}

\begin{table*}[!htbp]
\renewcommand\arraystretch{1.1}   
\caption{Performance comparison on USTC-CSL100.}
\vspace{-3mm}
\centering
\resizebox{\textwidth}{!}{
\begin{tabular}{lccccccccccccccc}
\toprule[1pt]
\multirow{2}{*}{Methods}&\multicolumn{7}{c}{Observation 20 Frames - Prediction 10 Frames} & ~ &\multicolumn{7}{c}{Observation 10 Frames - Prediction 20 Frames} \\
\cline{2-8}
\cline{10-16}
~ &B1$\uparrow$ &B4$\uparrow$ &ROUGE$\uparrow$ &WER$\downarrow$ &DTW$\downarrow$ &FID$\downarrow$ &MPJPE$\downarrow$& ~ &B1$\uparrow$ &B4$\uparrow$ &ROUGE$\uparrow$ &WER$\downarrow$ &DTW$\downarrow$ &FID$\downarrow$ &MPJPE$\downarrow$\\ 
\midrule[0.5pt]
\multicolumn{1}{l}{Ground Truth}  &69.10 &59.12 &68.53 &47.38 &0.00 &0.00 &0.00 &~ &69.10 &59.12 & 68.53&47.38 &0.00 &0.00 &0.00\\ 
\midrule[0.5pt]
\multicolumn{1}{l}{G2P-DDM~\cite{xie2024g2p}} &13.74 &2.11 &13.85 &92.04 &7.00 &0.53 &258.43 &~ &13.43 &1.74 &13.62 &92.15 &7.47 &0.53 &258.43 \\
\multicolumn{1}{l}{VQ-GCDM~\cite{tang2024GCDM}} &30.71 &10.68 &29.17 &77.87 &5.10 &0.43 &91.10 & ~ &18.77 &2.08 &16.64 &92.67 &10.89 &0.54 &104.27 \\
\midrule[0.5pt]
\multicolumn{1}{l}{\textbf{Ours}}
&\bf{69.64}&\bf{58.98}&\bf{69.26} &\bf{46.66}&\bf{1.92} &\bf{0.31} &\bf{77.04}& ~ &\bf{67.95}&\bf{56.61}&\bf{67.38} &\bf{49.24} &\bf{1.79} &\bf{0.31} &\bf{78.62}\\ 
\bottomrule[1pt]
\end{tabular}}
\label{tab:main2}
\vspace{-3mm}
\end{table*}

\subsection{Ablation Study}
In this subsection, we explore the effects of different random masking ratios in the training phase and various padding strategies during inference. The experiments in this part are all conducted under the condition of predicting 10 transition frames for every 20 observed sign frames to evaluate their impact on both semantic accuracy and transition coherence. 

\noindent\textbf{Effect of Masking Ratio.} 
Table~\ref{tab: random mask ratio 20-10} presents the results of different random masking ratios $r$ on semantic preservation and temporal coherence. With a lower masking ratio ($r$=0.1), the model achieves the best coherence, reflected by the lowest DTW scores of 0.61 on DEV and 0.60 on TEST. This indicates that smaller masking ratios allow the model to maintain smoother transitions, as more contextual frames are available to guide the generation of missing frames. However, as the masking ratio increases to $r$=0.3, we observe a slight improvement in semantic accuracy, with BLEU-4 scores rising to 8.06 on DEV and 8.73 on TEST. This suggests that a higher masking ratio may encourage the model to focus more on the sign semantic content, as it learns to reconstruct more extensive missing segments. 

Interestingly, further increasing the masking ratio to $r$=0.5 leads to a drop in both semantic accuracy and coherence, with DTW scores increasing and BLEU scores decreasing slightly. This highlights a trade-off: while moderate masking ratios can improve semantic preservation, excessive masking may reduce the model's ability to maintain smooth transitions and semantic accuracy. Based on these observations, we select $r$=0.3 for subsequent experiments, as it provides a good balance between semantic retention and transition smoothness. 



\noindent\textbf{Influence of Padding Strategy.} 
Table~\ref{tab: ablation padding} summarizes the results of different padding strategies applied during inference. We experiment with four approaches: w/o Padding (no initialization for missing frames), Front Padding (using the last frame of the previous observation segment), Back Padding (using the first frame of the following observation segment), and Linear Padding (performing linear interpolation between the two boundary poses, as Eq.~\ref{eq:padding}). 
The “w/o Padding” approach yields the highest DTW scores (11.46 on DEV and 10.36 on TEST), indicating poor coherence and unnatural transitions. Both Front Padding and Back Padding improve the smoothness of transitions, as shown by lower DTW scores (0.67 on DEV and 0.66 on TEST), but their FID and MPJPE scores remain higher compared to Linear Padding. 

The Linear Padding strategy achieves the best overall performance, with the lowest DTW scores (0.62 on DEV and 0.66 on TEST) and MPJPE values, reflecting both smoothness and accuracy in pose reconstruction. By initializing missing frames through linear interpolation, Linear Padding provides a stable and effective foundation for generating realistic and seamless transitions, closely resembling real-world signing behavior.

\begin{figure*}[t]
\centering
\includegraphics[width=\textwidth]{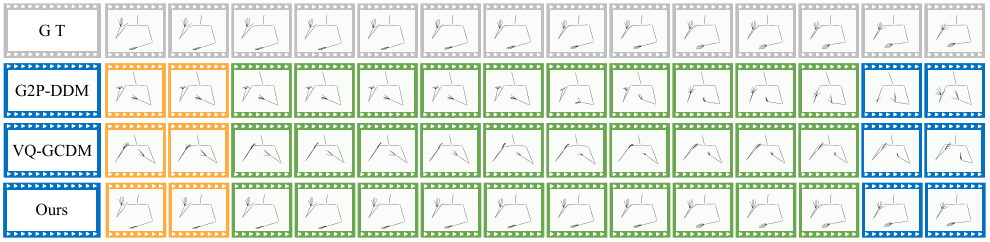}
\vspace{-8mm}
\caption{Visualization examples of generating 10-frame transition pose under 20-frame observations on PHOENIX14T. We compare our method with G2P-DDM and VQ-GCDM, attached the Ground Truth. \textcolor{green}{Green}: Transitions. \textcolor{orange}{Orange} and \textcolor{blue}{Blue}: Observations.} 
\label{fig: PHOENIX14T 20-10}
\vspace{-3mm}
\end{figure*}

\begin{figure*}[t]
\centering
\includegraphics[width=\textwidth]{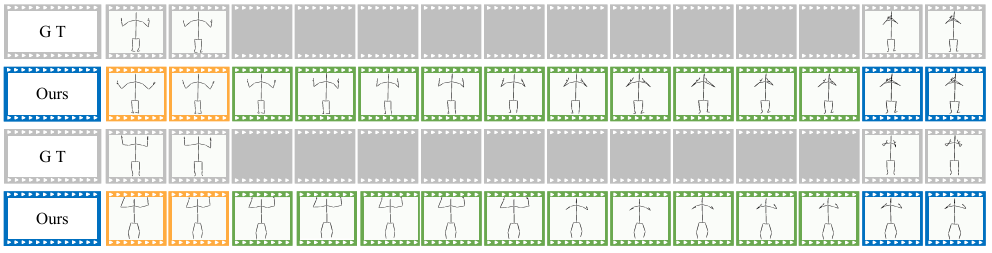}
\vspace{-8mm}
\caption{Visualization examples on the USTC-SLR500, demonstrate the generation of transition poses according to two discrete sign segments. \textcolor{green}{Green}: Transitions. \textcolor{orange}{Orange} and \textcolor{blue}{Blue}: Observations.}
\label{fig: USTC-SLR500}
\vspace{-3mm}
\end{figure*}

\subsection{Comparison with State-of-the-Arts}
We compare our method with two state-of-the-art models: G2P-DDM~\cite{xie2024g2p} and VQ-GCDM~\cite{tang2024GCDM}. Both G2P-DDM and VQ-GCDM are primarily designed for SLP. Due to the absence of established benchmarks for our specific task, we adapted these models to serve as comparative baselines.

\noindent\textbf{Comparison on PHOENIX14T.}
As shown in Table~\ref{tab:main1}, under the “Observation 20 Frames - Prediction 10 Frames” setting, our method achieves superior performance across both semantic and coherence metrics. Specifically, we attain a BLEU-1 score of 22.46\% and 22.36\% on the DEV and TEST sets, respectively, outperforming G2P-DDM and VQ-GCDM. This improvement in BLEU scores indicates that our method more accurately preserves semantic content compared to existing methods. Furthermore, our method shows significant gains in temporal coherence, with DTW scores of 0.62 on DEV and 0.61 on TEST, respectively. These results confirm that our model generates smoother and more realistic transitions than the baseline methods.

Under the more challenging “Observation 10 Frames - Prediction 20 Frames” setting, our method continues to demonstrate robust performance despite the increased prediction length. We achieve BLEU-1 scores of 19.32\% on DEV and 19.59\% on TEST, considerably higher than the baseline methods. Our method also maintains lower DTW and FID scores, demonstrating its ability to generate coherent and semantically accurate sequences even with limited observed data. The MPJPE scores further support the effectiveness of our approach, with scores of 48.04 on DEV and 59.81 on TEST, indicating a closer alignment to ground truth poses than the other methods.

\noindent\textbf{Comparison on USTC-CSL100.}
Table~\ref{tab:main2} shows the results on the USTC-CSL100 dataset, which evaluates our method’s ability to generate sign language sequences. In the “Observation 20 Frames - Prediction 10 Frames” setting, our model achieves near-ground-truth semantic accuracy, with BLEU-1 and BLEU-4 scores of 69.64\% and 58.98\% on DEV, closely approaching the ground truth scores of 69.10\% and 59.12\%. Additionally, our DTW score of 1.922 and FID score of 3.21 indicate high temporal consistency and low distributional divergence from the real data. Compared to G2P-DDM and VQ-GCDM, which yield higher DTW and MPJPE scores, our model achieves smoother and more accurate pose transitions.

In the “Observation 10 Frames - Prediction 20 Frames” scenario, our method continues to demonstrate resilience in generating long sequences. Our BLEU-1 and BLEU-4 scores on DEV are 67.95\% and 56.61\%, respectively, showing superior semantic preservation compared to the baseline methods. Additionally, the DTW score of 1.79 and MPJPE score of 78.62 confirm our model's ability to produce coherent and realistic transitions, outperforming the baselines in alignment with true pose sequences. 




\subsection{Qualitative Results}

\noindent\textbf{Visualization Results.} Figure~\ref{fig: PHOENIX14T 20-10} presents a comparative demonstration of our method against other approaches (\emph{i.e.}, G2P-DDM~\cite{xie2024g2p} and VQ-GCDM~\cite{tang2024GCDM}) in generating transition sign poses. This example clearly shows that our method significantly outperforms G2P-DDM and VQ-GCDM in the quality of interpolated movements, especially excelling in overall motion coherence. Furthermore, our method meticulously fine-tunes the observations (the \textcolor{blue}{Blue} and \textcolor{orange}{Orange} sections), effectively avoiding coherence issues that may arise from directly concatenating discrete sign segments.

\noindent\textbf{Verification in Realistic Scenarios.} Figure~\ref{fig: USTC-SLR500} demonstrates visualization examples for synthesizing continuous sign videos from discrete segments in a realistic, word-level scenario using the USTC-SLR500 dataset. Here, the observed frames from two isolated sign segments are marked in blue and orange, while the generated transitions are shown in green. The results show that our method not only generates smooth transitions between discrete sign language segments but also fine-tunes the observed poses based on the overall semantics, resulting in a coherent and continuous sign language video.
\section{Conclusions}
\label{sec:conclusions}
This paper proposes a method to generate continuous sign language videos by synthesizing transition frames between discrete segments, transforming an unsupervised task into a supervised one through random masking. Using a conditional diffusion model, the approach learns to predict missing frames based on observable context, allowing for smooth and temporally coherent outputs. By incorporating a linearly interpolating padding strategy during inference, the model initializes transitions effectively, further refining them into seamless sequences. Extensive experiments demonstrate the effectiveness of leveraging context for natural transitions, advancing sign language synthesis by enabling fluid, continuous sequences. 

{
    \small
    \bibliographystyle{ieeenat_fullname}
    \bibliography{main}
}


\end{document}